\documentclass{article}
\pdfoutput=1 

\usepackage{amsmath, amssymb}           
\usepackage{graphicx}                   
\usepackage{pdfpages}
\usepackage{xcolor}
\usepackage{color}
\usepackage[sc]{titlesec}
\usepackage{natbib}
\usepackage{iclr2017_conference,times}
\usepackage{dsfont}
\usepackage{float}
\usepackage{cleveref}
\usepackage{tabularx}
\newcolumntype{Y}{>{\centering\arraybackslash}X}
\newcolumntype{s}{>{\hsize=.5\hsize}Y}
\newcolumntype{b}{X}
\restylefloat{table}
\usepackage[breaklinks=true,colorlinks,citecolor=black,bookmarks=false]{hyperref}
\hypersetup{
  pdfinfo={
    Title={A Baseline for Detecting Misclassified and Out-of-Distribution Examples in Neural Networks},
    Author={Dan Hendrycks and Kevin Gimpel},
    Subject={Deep Learning, Neural Networks, Anomaly Detection},
    Keywords={out-of-distribution, OOD, anomaly, outlier, out of distribution, confidence, calibration, anomaly detection, calibration, outlier detection, open category, out of sample, out-of-sample, safety, ai safety, deep networks, machine learning, neural networks, image classifiers, convnets, convolutional neural networks}
  }
}

\iclrfinalcopy

\title{A Baseline for Detecting Misclassified and Out-of-Distribution Examples\\ in Neural Networks}

\author{Dan Hendrycks\thanks{Work done while the author was at TTIC. Code is available at \href{https://github.com/hendrycks/error-detection
}{github.com/hendrycks/error-detection
} }\\University of California, Berkeley\\\texttt{hendrycks@berkeley.edu} \And Kevin Gimpel\\ Toyota Technological Institute at Chicago\\\texttt{kgimpel@ttic.edu}}

\begin{document}

\date{}
\maketitle
\begin{abstract}
We consider the two related problems of detecting if an example is misclassified or out-of-distribution. We present a simple baseline that utilizes probabilities from softmax distributions. Correctly classified examples tend to have greater maximum softmax probabilities than erroneously classified and out-of-distribution examples, allowing for their detection. We assess performance by defining several tasks in computer vision, natural language processing, and automatic speech recognition, showing the effectiveness of this baseline across all. We then show the baseline can sometimes be surpassed, demonstrating the room for future research on these underexplored detection tasks.
\end{abstract}

\section{Introduction}
When machine learning classifiers are employed in real-world tasks, they tend to fail when the training and test distributions differ. Worse, these classifiers often fail silently by providing high-confidence predictions while being woefully incorrect \citep{goodfellow, colah}. Classifiers failing to indicate when they are likely mistaken can limit their adoption or cause serious accidents. For example, a medical diagnosis model may consistently classify with high confidence, even while it should flag difficult examples for human intervention. The resulting unflagged, erroneous diagnoses could blockade future machine learning technologies in medicine. More generally and importantly, estimating when a model is in error is of great concern to AI Safety \citep{colah}.

These high-confidence predictions are frequently produced by softmaxes because softmax probabilities are computed with the fast-growing exponential function. Thus minor additions to the softmax inputs, i.e. the logits, can lead to substantial changes in the output distribution. Since the softmax function is a smooth approximation of an indicator function, it is uncommon to see a uniform distribution outputted for out-of-distribution examples. Indeed, random Gaussian noise fed into an MNIST image classifier gives a ``prediction confidence'' or predicted class probability of 91\%, as we show later. Throughout our experiments we establish that the prediction probability from a softmax distribution has a poor direct correspondence to confidence. This is consistent with a great deal of anecdotal evidence from researchers \citep{nguyen,yu,provost,foolers}.

However, in this work we also show the prediction probability of incorrect and out-of-distribution examples tends to be lower than the prediction probability for correct examples. Therefore, capturing prediction probability statistics about correct or in-sample examples is often sufficient for detecting whether an example is in error or abnormal, even though the prediction probability viewed in isolation can be misleading.

These prediction probabilities form our detection baseline, and we demonstrate its efficacy through various computer vision, natural language processing, and automatic speech recognition tasks. While these prediction probabilities create a consistently useful baseline, at times they are less effective, revealing room for improvement. To give ideas for future detection research, we 
contribute one method which outperforms the baseline on some (but not all) tasks. This new method evaluates the quality of a neural network's input reconstruction to determine if an example is abnormal.

In addition to the baseline methods, another contribution of this work is the designation of standard tasks and evaluation metrics for assessing the automatic detection of errors and out-of-distribution examples. We use a large number of well-studied tasks across three research areas, using standard neural network architectures that perform well on them. For out-of-distribution detection, we provide ways to supply the out-of-distribution examples at test time like using images from different datasets and realistically distorting inputs. 
We hope that other researchers will pursue these tasks in future work and surpass the performance of our baselines. 

In summary, while softmax classifier probabilities are not directly useful as confidence estimates, estimating model confidence is not as bleak as previously believed. Simple statistics derived from softmax distributions provide a surprisingly effective way to determine whether an example is misclassified or from a different distribution from the training data, as demonstrated by our experimental results spanning computer vision, natural language processing, and speech recognition tasks. This creates a strong baseline for detecting errors and out-of-distribution examples which we hope future research surpasses. 

\section{Problem Formulation and Evaluation}
In this paper, we are interested in two related problems. The first is \textbf{error and success prediction}: can we predict whether a trained classifier will make an error on a particular held-out test example; can we predict if it will correctly classify said example? The second is \textbf{in- and out-of-distribution detection}: can we predict whether a test example is from a different distribution from the training data; can we predict if it is from within the same distribution?\footnote{We consider adversarial example detection techniques in a separate work \citep{detectadversarial}.} Below we present a simple baseline for solving these two problems. To evaluate our solution, we use two evaluation metrics.

Before mentioning the two evaluation metrics, we first note that comparing detectors is not as straightforward as using accuracy. For detection we have two classes, and the detector outputs a score for both the positive and negative class. If the negative class is far more likely than the positive class, a model may always guess the negative class and obtain high accuracy, which can be misleading \citep{provost}. We must then specify a score threshold so that some positive examples are classified correctly, but this depends upon the trade-off between false negatives (fn) and false positives (fp).

Faced with this issue, we employ the Area Under the Receiver Operating Characteristic curve (AUROC) metric, which is a threshold-independent performance evaluation \citep{auroc}. The ROC curve is a graph showing the true positive rate ($\text{tpr}=\text{tp}/(\text{tp} + \text{fn})$) and the false positive rate ($\text{fpr}=\text{fp}/(\text{fp} + \text{tn})$) against each other. Moreover, the AUROC can be interpreted as the probability that a positive example has a greater detector score/value than a negative example \citep{ROC}. Consequently, a random positive example detector corresponds to a 50\% AUROC, and a ``perfect'' classifier corresponds to 100\%.\footnote{A debatable, imprecise interpretation of AUROC values may be as follows: 90\%|100\%: Excellent, \; 80\%|90\%: Good, \; 70\%|80\%: Fair, \; 60\%|70\%: Poor, \; 50\%|60\%: Fail.}

The AUROC sidesteps the issue of threshold selection, as does the Area Under the Precision-Recall curve (AUPR) which is sometimes deemed more informative \citep{manning}. This is because the AUROC is not ideal when the positive class and negative class have greatly differing base rates, and the AUPR adjusts for these different positive and negative base rates. For this reason, the AUPR is our second evaluation metric. The PR curve plots the precision ($\text{tp}/(\text{tp} + \text{fp})$) and recall ($\text{tp}/(\text{tp} + \text{fn})$) against each other. The baseline detector has an AUPR approximately equal to the precision \citep{auprbaseline}, and a ``perfect'' classifier has an AUPR of $100\%$. Consequently, the base rate of the positive class greatly influences the AUPR, so for detection we must specify which class is positive. In view of this, we show the AUPRs when we treat success/normal classes as positive, and then we show the areas when we treat the error/abnormal classes as positive. We can treat the error/abnormal classes as positive by multiplying the scores by $-1$ and labeling them positive. Note that treating error/abnormal classes as positive classes does not change the AUROC since if $S$ is a score for a  successfully classified value, and $E$ is the score for an erroneously classified value, $\text{AUROC}=P(S>E) = P(-E>-S)$.
%

We begin our experiments in Section~\ref{sec:softmax-stats} where we describe a simple baseline which uses the maximum probability from the softmax label distribution in neural network classifiers. Then in Section~\ref{sec:auxiliary} we describe a method that uses an additional, auxiliary model component trained to reconstruct the input.

\section{Softmax Prediction Probability as a Baseline}
\label{sec:softmax-stats}
In what follows we retrieve the maximum/predicted class probability from a softmax distribution and thereby detect whether an example is erroneously classified or out-of-distribution. Specifically, we separate correctly and incorrectly classified \emph{test set} examples and, for each example, compute the softmax probability of the predicted class, i.e., the maximum softmax probability.\footnote{We also tried using the KL divergence of the softmax distribution from the uniform distribution for detection. With divergence values, detector AUROCs and AUPRs were highly correlated with AUROCs and AUPRs from a detector using the maximum softmax probability. This divergence is similar to entropy \citep{jacob, williams}.} From these two groups we obtain the area under PR and ROC curves. These areas summarize the performance of a binary classifier discriminating with values/scores (in this case, maximum probabilities from the softmaxes) across different thresholds. This description treats correctly classified examples as the positive class, denoted ``Success'' or ``Succ'' in our tables. In ``Error'' or ``Err'' we treat the the incorrectly classified examples as the positive class; to do this we label incorrectly classified examples as positive and take the negatives of the softmax probabilities of the predicted classes as the scores.

For ``In,'' we treat the in-distribution, correctly classified test set examples as positive and use the softmax probability for the predicted class as a score, while for ``Out'' we treat the out-of-distribution examples as positive and use the negative of the aforementioned probability. Since the AUPRs for Success, Error, In, Out classifiers depend on the rate of positive examples, we list what area a random detector would achieve with ``Base'' values. Also in the upcoming results we list the mean predicted class probability of wrongly classified examples (Pred Prob Wrong (mean)) to demonstrate that the softmax prediction probability is a misleading confidence proxy when viewed in isolation. The ``Pred. Prob (mean)'' columns show this same shortcoming but for out-of-distribution examples.

Table labels aside, we begin experimentation with datasets from vision then consider tasks in natural language processing and automatic speech recognition. In all of the following experiments, the AUROCs differ from the random baselines with high statistical significance according to the Wilcoxon rank-sum test.

\subsection{Computer Vision}
In the following computer vision tasks, we use three datasets: MNIST, CIFAR-10, and CIFAR-100 \citep{cifar}. MNIST is a dataset of handwritten digits, consisting of 60000 training and 10000 testing examples. Meanwhile, CIFAR-10 has colored images belonging to 10 different classes, with 50000 training and 10000 testing examples. CIFAR-100 is more difficult, as it has 100 different classes with 50000 training and 10000 testing examples.

In Table \ref{tab:visionerrors}, we see that correctly classified and incorrectly classified examples are sufficiently distinct and thus allow reliable discrimination. Note that the area under the curves degrade with image recognizer test error.

\begin{table}
\begin{center}
\begin{tabularx}{\textwidth}{X | *{5}{>{\hsize=.8\hsize}Y}}
\hline Dataset & AUROC\newline/Base & AUPR Succ/Base & AUPR Err/Base & Pred.~Prob Wrong(mean) & Test Set Error\\ \cline{1-6}
\bf{MNIST} 		& 97\scalebox{1.}{/50} & 100\scalebox{1.}{/98} & 48\scalebox{1.}{/1.7}  & 86 & 1.69 \\
\bf{CIFAR-10} 	& 93\scalebox{1.}{/50} & 100\scalebox{1.}{/95} & 43\scalebox{1.}{/5} & 80 & 4.96 \\
\bf{CIFAR-100} 	& 87\scalebox{1.}{/50} & 96\scalebox{1.}{/79} & 62\scalebox{1.}{/21} & 66 & 20.7 \\
\hline
\end{tabularx}
\caption{The softmax predicted class probability allows for discrimination between correctly and incorrectly classified test set examples. ``Pred.~Prob Wrong(mean)'' is the mean softmax probability for wrongly classified examples, showcasing its shortcoming as a direct measure of confidence. Succ/Err Base values are the AUROCs or AUPRs achieved by random classifiers. All entries are percentages.}\label{tab:visionerrors}
\end{center}
\end{table}

Next, let us consider using softmax distributions to determine whether an example is in- or out-of-distribution. We use all test set examples as the in-distribution (positive) examples. For out-of-distribution (negative) examples, we use realistic images and noise. For CIFAR-10 and CIFAR-100, we use realistic images 
from the Scene UNderstanding dataset (SUN), which consists of 397 different scenes \citep{SUN}. For MNIST, we use grayscale realistic images from three sources. 
Omniglot~\citep{omniglot} images are handwritten characters rather than the handwritten digits in MNIST. 
Next, notMNIST~\citep{notmnist} consists of typeface characters. Last of the realistic images, CIFAR-10bw are black and white rescaled CIFAR-10 images. The synthetic ``Gaussian'' data is random normal noise, and ``Uniform'' data is random uniform noise. Images are resized when necessary.

The results are shown in Table~\ref{tab:visionood}. Notice that the mean predicted/maximum class probabilities (Pred. Prob (mean)) are above 75\%, 
but if the prediction probability alone is translated to confidence, the softmax distribution should be more uniform for CIFAR-100. This again shows softmax probabilities should not be viewed as a direct representation of confidence. Fortunately, out-of-distribution examples sufficiently differ in the prediction probabilities from in-distribution examples, allowing for successful detection and generally high area under PR and ROC curves.

\begin{table}
\begin{center}
\begin{tabularx}{\textwidth}{X | *{4}{>{\hsize=.45\hsize}Y}}
\hline In-Distribution /\newline Out-of-Distribution & AUROC\newline/Base & AUPR In\newline/Base & AUPR Out/Base & Pred. Prob (mean) \\ \cline{1-5}
\bf{CIFAR-10/SUN} 		& 95\scalebox{1.}{/50}  & 89\scalebox{1.}{/33} & 97\scalebox{1.}{/67} & 72 \\
\bf{CIFAR-10/Gaussian}	& 97\scalebox{1.}{/50}  & 98\scalebox{1.}{/49} & 95\scalebox{1.}{/51} & 77 \\
\bf{CIFAR-10/All}     & 96\scalebox{1.}{/50}  & 88\scalebox{1.}{/24} & 98\scalebox{1.}{/76} & 74 \\
\hline
\bf{CIFAR-100/SUN}		& 91\scalebox{1.}{/50}  & 83\scalebox{1.}{/27} & 96\scalebox{1.}{/73} & 56 \\
\bf{CIFAR-100/Gaussian} & 88\scalebox{1.}{/50}  & 92\scalebox{1.}{/43} & 80\scalebox{1.}{/57} & 77 \\
\bf{CIFAR-100/All}     & 90\scalebox{1.}{/50}  & 81\scalebox{1.}{/21} & 96\scalebox{1.}{/79} & 63 \\
\hline
\bf{MNIST/Omniglot}		& 96\scalebox{1.}{/50}  & 97\scalebox{1.}{/52} & 96\scalebox{1.}{/48} & 86 \\
\bf{MNIST/notMNIST}		& 85\scalebox{1.}{/50}  & 86\scalebox{1.}{/50} & 88\scalebox{1.}{/50} & 92 \\
\bf{MNIST/CIFAR-10bw}	& 95\scalebox{1.}{/50}  & 95\scalebox{1.}{/50} & 95\scalebox{1.}{/50} & 87 \\
\bf{MNIST/Gaussian}		& 90\scalebox{1.}{/50}  & 90\scalebox{1.}{/50} & 91\scalebox{1.}{/50} & 91 \\
\bf{MNIST/Uniform}		& 99\scalebox{1.}{/50}  & 99\scalebox{1.}{/50} & 98\scalebox{1.}{/50} & 83 \\
\bf{MNIST/All}         & 91\scalebox{1.}{/50}  & 76\scalebox{1.}{/20} & 98\scalebox{1.}{/80} & 89 \\
\hline
\end{tabularx}
\caption{Distinguishing in- and out-of-distribution test set data for image classification. CIFAR-10/All is the same as CIFAR-10/(SUN, Gaussian). All values are percentages.}\label{tab:visionood}
\end{center}
\end{table}

For reproducibility, let us specify the model architectures. The MNIST classifier is a three-layer, 256 neuron-wide, fully-connected network trained for 30 epochs with Adam \citep{adam}. It uses a GELU nonlinearity \citep{gelu}, $x\Phi(x)$, where $\Phi(x)$ is the CDF of the standard normal distribution. We initialize our weights according to \citep{init}, as it is suited for arbitrary nonlinearities. For CIFAR-10 and CIFAR-100, we train a 40-4 wide residual network \citep{wrn} for 50 epochs with stochastic gradient descent using restarts \citep{sgdr}, the GELU nonlinearity, and standard mirroring and cropping data augmentation.

\subsection{Natural Language Processing}
Let us turn to a variety of tasks and architectures used in natural language processing.
\subsubsection{Sentiment Classification}
The first NLP task is binary sentiment classification using the IMDB dataset \citep{maas}, a dataset of polarized movie reviews with 25000 training and 25000 test reviews. This task allows us to determine if classifiers trained on a relatively small dataset still produce informative softmax distributions. For this task we use a linear classifier taking as input the average of trainable, randomly initialized word vectors with dimension 50 \citep{fasttext, DAN}. We train for 15 epochs with Adam and early stopping based upon 5000 held-out training reviews. Again, Table \ref{tab:imdberr} shows that the softmax distributions differ between correctly and incorrectly classified examples, so prediction probabilities allow us to detect reliably which examples are right and wrong.

Now we use the Customer Review \citep{CR} and Movie Review \citep{MR} datasets as out-of-distribution examples. The Customer Review dataset has reviews of products rather than only movies, and the Movie Review dataset has snippets from professional movie reviewers rather than full-length amateur reviews. We leave all test set examples from IMDB as in-distribution examples, and out-of-distribution examples are the 500 or 1000 test reviews from Customer Review and Movie Review datasets, respectively. Table \ref{tab:imdbood} displays detection results, showing a similar story to Table 2.

\begin{table}
\begin{center}
\begin{tabularx}{\textwidth}{X | *{5}{>{\hsize=1\hsize}Y}}
\hline Dataset & AUROC\newline/Base & AUPR Succ/Base & AUPR Err/Base & Pred. Prob Wrong(mean) & Test Set Error\\ \cline{1-6}
\bf{IMDB}   & 82\scalebox{1.}{/50} & 97\scalebox{1.}{/88} & 36\scalebox{1.}{/12} & 74 & 11.9 \\
\hline
\end{tabularx}
\caption{Detecting correct and incorrect classifications for binary sentiment classification. 
}\label{tab:imdberr}
\end{center}
\end{table}

\begin{table}
\begin{center}
\begin{tabularx}{\textwidth}{X | *{4}{>{\hsize=.4\hsize}Y}}
\hline In-Distribution /\newline Out-of-Distribution & AUROC\newline/Base & AUPR In\newline/Base & AUPR Out/Base & Pred. Prob (mean) \\ \cline{1-5}
\bf{IMDB/Customer Reviews} 	& 95\scalebox{1.}{/50} & 99\scalebox{1.}{/89} & 60\scalebox{1.}{/11} & 62 \\
\bf{IMDB/Movie Reviews}		& 94\scalebox{1.}{/50} & 98\scalebox{1.}{/72} & 80\scalebox{1.}{/28} & 63 \\
\bf{IMDB/All}		& 94\scalebox{1.}{/50} & 97\scalebox{1.}{/66} & 84\scalebox{1.}{/34} & 63 \\
\hline
\end{tabularx}
\caption{Distinguishing in- and out-of-distribution test set data for binary sentiment classification. IMDB/All is the same as IMDB/(Customer Reviews, Movie Reviews). All values are percentages.}\label{tab:imdbood}
\end{center}
\end{table}

\subsubsection{Text Categorization}
We turn to text categorization tasks to determine whether softmax distributions are useful for detecting similar but out-of-distribution examples. In the following text categorization tasks, we train classifiers to predict the subject of the text they are processing. In the 20 Newsgroups dataset \citep{newsgroups}, there are 20 different newsgroup subjects with a total of 20000 documents for the whole dataset. The Reuters 8 \citep{reuters} dataset has eight different news subjects with nearly 8000 stories in total. The Reuters 52 dataset has 52 news subjects with slightly over 9000 news stories; this dataset can have as few as three stories for a single subject. 

For the 20 Newsgroups dataset we train a linear classifier on 30-dimensional word vectors for 20 epochs. Meanwhile, Reuters 8 and Retuers 52 use one-layer neural networks with a bag-of-words input and a GELU nonlinearity, all optimized with Adam for 5 epochs. We train on a subset of subjects, leaving out 5 newsgroup subjects from 20 Newsgroups, 2 news subjects from Reuters 8, and 12 news subjects from Reuters 52, leaving the rest as out-of-distribution examples. Table \ref{tab:caterr} shows that with these datasets and architectures, we can detect errors dependably, and Table \ref{tab:catood} informs us that the softmax prediction probabilities allow for detecting out-of-distribution subjects.

\begin{table}
\begin{center}
\begin{tabularx}{\textwidth}{X | *{5}{>{\hsize=.6\hsize}Y}}
\hline Dataset & AUROC\newline/Base & AUPR Succ/Base & AUPR Err/Base & Pred.Prob Wrong(mean) & Test Set Error\\ \cline{1-6}
\bf{15 Newsgroups}   	& 89\scalebox{1.}{/50} & 99\scalebox{1.}{/93} & 42\scalebox{1.}{/7.3} & 53 & 7.31 \\
\bf{Reuters 6}   		& 89\scalebox{1.}{/50} & 100\scalebox{1.}{/98} & 35\scalebox{1.}{/2.5} & 77 & 2.53 \\
\bf{Reuters 40}   		& 91\scalebox{1.}{/50} & 99\scalebox{1.}{/92} & 45\scalebox{1.}{/7.6} & 62 & 7.55 \\
\hline
\end{tabularx}
\caption{Detecting correct and incorrect classifications for text categorization. 
}\label{tab:caterr}
\end{center}
\end{table}

\begin{table}
\begin{center}
\begin{tabularx}{\textwidth}{X | *{4}{>{\hsize=.5\hsize}Y}}
\hline In-Distribution /\newline Out-of-Distribution & AUROC\newline/Base & AUPR In/Base & AUPR Out/Base & Pred. Prob (mean) \\ \cline{1-5}
\bf{15/5 Newsgroups}		& 75\scalebox{1.}{/50} & 92\scalebox{1.}{/84} & 45\scalebox{1.}{/16} & 65 \\
\bf{Reuters6/Reuters2}		& 92\scalebox{1.}{/50} & 100\scalebox{1.}{/95} & 56\scalebox{1.}{/4.5} & 72 \\
\bf{Reuters40/Reuters12}	& 95\scalebox{1.}{/50} & 100\scalebox{1.}{/93} & 60\scalebox{1.}{/7.2} & 47 \\
\hline
\end{tabularx}
\caption{Distinguishing in- and out-of-distribution test set data for text categorization. 
}\label{tab:catood}
\end{center}
\end{table}

\subsubsection{Part-of-Speech Tagging}
Part-of-speech (POS) tagging of newswire and social media text 
is our next challenge. We use the Wall Street Journal portion of the Penn Treebank~\citep{WSJ} which contains 45 distinct POS tags. For social media, we use POS-annotated tweets~\citep{gimpel,owoputi} which contain 25 tags.
For the WSJ tagger, we train a bidirectional long short-term memory recurrent neural network~\citep{lstm} with three layers, 128 neurons per layer, with randomly initialized word vectors, and this is trained on $90\%$ of the corpus for 10 epochs with stochastic gradient descent with a batch size of 32. The tweet tagger is simpler, as it is two-layer neural network with a GELU nonlinearity, a weight initialization according to \citep{init}, pretrained word vectors trained on a corpus of 56 million tweets~\citep{owoputi}, and a hidden layer size of 256, all while training on 1000 tweets for 30 epochs with Adam and early stopping with 327 validation tweets. Error detection results are in Table \ref{tab:poserr}. For out-of-distribution detection, we use the WSJ tagger on the tweets as well as weblog data from the English Web Treebank \citep{engwebtb}. The results are shown in Table \ref{tab:posood}. Since the weblog data is closer in style to newswire than are the tweets, 
it is harder to detect whether a weblog sentence is out-of-distribution than a tweet. Indeed, since POS tagging is done at the word-level, we are detecting whether each word is out-of-distribution given the word and contextual features. With this in mind, we see that it is easier to detect words as out-of-distribution if they are from tweets than from blogs.

\begin{table}
\begin{center}
\begin{tabularx}{\textwidth}{X | *{5}{>{\hsize=1\hsize}Y}}
\hline Dataset & AUROC\newline/Base & AUPR Succ/Base & AUPR Err/Base & Pred. Prob Wrong(mean) & Test Set Error\\ \cline{1-6}
\bf{WSJ}   		& 96\scalebox{1.}{/50} & 100\scalebox{1.}{/96} & 51\scalebox{1.}{/3.7} & 71 & 3.68 \\
\bf{Twitter}  	& 89\scalebox{1.}{/50} & 98\scalebox{1.}{/87} & 53\scalebox{1.}{/13} & 69 & 12.59 \\
\hline
\end{tabularx}
\caption{Detecting correct and incorrect classifications for part-of-speech tagging. 
}\label{tab:poserr}
\end{center}
\end{table}

\begin{table}[H]
\begin{center}
\begin{tabularx}{\textwidth}{X | *{4}{>{\hsize=.6\hsize}Y}}
\hline In-Distribution /\newline Out-of-Distribution & AUROC\newline/Base & AUPR In/Base & AUPR Out/Base & Pred. Prob (mean) \\ \cline{1-5}
\bf{WSJ/Twitter}	& 80\scalebox{1.}{/50} & 98\scalebox{1.}{/92} & 41\scalebox{1.}{/7.7} & 81 \\
\bf{WSJ/Weblog*}	& 61\scalebox{1.}{/50} & 88\scalebox{1.}{/86} & 30\scalebox{1.}{/14} & 93 \\
\hline
\end{tabularx}
\caption{Detecting out-of-distribution tweets and blog articles for part-of-speech tagging. All values are percentages. *These examples are atypically close to the training distribution.}\label{tab:posood}
\end{center}
\end{table}

\subsection{Automatic Speech Recognition}
Now we consider a task which uses softmax values to construct entire sequences rather than determine an input's class. 
Our sequence prediction system uses a bidirectional LSTM with two-layers and a clipped GELU nonlinearity, optimized for 60 epochs with RMSProp trained on $80\%$ of the TIMIT corpus \citep{timit}. The LSTM is trained with connectionist temporal classification (CTC) \citep{ctc} for predicting sequences of phones given MFCCs, energy, and first and second deltas of a 25ms frame. When trained with CTC, the LSTM learns to have its phone label probabilities spike momentarily while mostly predicting blank symbols otherwise. In this way, the softmax is used differently from typical classification problems, providing a unique test for our detection methods. 

We do not show how the system performs on correctness/incorrectness detection because errors are not binary and instead lie along a range of edit distances. However, we can perform out-of-distribution detection. Mixing the TIMIT audio with realistic noises from the Aurora-2 dataset \citep{aurora}, we keep the TIMIT audio volume at 100\% and noise volume at 30\%, giving a mean SNR of approximately 5. Speakers are still clearly audible to the human ear but confuse the phone recognizer because the prediction edit distance more than doubles. For more out-of-distribution examples, we use the test examples from the THCHS-30 dataset \citep{chinese}, a Chinese speech corpus. Table \ref{tab:ctctimitood} shows the results. Crucially, when performing detection, we compute the softmax probabilities while ignoring the blank symbol's logit. With the blank symbol's presence, the softmax distributions at most time steps predict a blank symbol with high confidence, but without the blank symbol we can better differentiate between normal and abnormal distributions. With this modification, the softmax prediction probabilities allow us to detect whether an example is out-of-distribution.

\begin{table}
\begin{center}
\begin{tabularx}{\textwidth}{X | *{4}{>{\hsize=.4\hsize}Y}}
\hline In-Distribution /\newline Out-of-Distribution & AUROC\newline/Base & AUPR In/Base & AUPR Out/Base & Pred. Prob (mean)\\ \cline{1-5}
\bf{TIMIT/TIMIT+Airport}		& 99\scalebox{1.}{/50} & 99\scalebox{1.}{/50} & 99\scalebox{1.}{/50} & 59 \\
\bf{TIMIT/TIMIT+Babble}		& 100\scalebox{1.}{/50} & 100\scalebox{1.}{/50} & 100\scalebox{1.}{/50} & 55 \\
\bf{TIMIT/TIMIT+Car}			& 98\scalebox{1.}{/50} & 98\scalebox{1.}{/50} & 98\scalebox{1.}{/50} & 59 \\
\bf{TIMIT/TIMIT+Exhibition}	& 100\scalebox{1.}{/50} & 100\scalebox{1.}{/50} & 100\scalebox{1.}{/50} & 57 \\
\bf{TIMIT/TIMIT+Restaurant}	& 98\scalebox{1.}{/50} & 98\scalebox{1.}{/50} & 98\scalebox{1.}{/50} & 60 \\
\bf{TIMIT/TIMIT+Street}		& 100\scalebox{1.}{/50} & 100\scalebox{1.}{/50} & 100\scalebox{1.}{/50} & 52 \\
\bf{TIMIT/TIMIT+Subway}		& 100\scalebox{1.}{/50} & 100\scalebox{1.}{/50} & 100\scalebox{1.}{/50} & 56 \\
\bf{TIMIT/TIMIT+Train}		& 100\scalebox{1.}{/50} & 100\scalebox{1.}{/50} & 100\scalebox{1.}{/50} & 58 \\
\bf{TIMIT/Chinese}		& 85\scalebox{1.}{/50} & 80\scalebox{1.}{/34} & 90\scalebox{1.}{/66} & 64 \\
\bf{TIMIT/All}		& 97\scalebox{1.}{/50} & 79\scalebox{1.}{/10} & 100\scalebox{1.}{/90} & 58 \\
\hline
\end{tabularx}
\caption{Detecting out-of-distribution distorted speech. All values are percentages.}\label{tab:ctctimitood}
\end{center}
\end{table}

\section{Abnormality Detection with Auxiliary Decoders}
\label{sec:auxiliary}
\begin{figure}[H]
	\centering
	\noindent\makebox[\textwidth]{\includegraphics[scale=0.52]{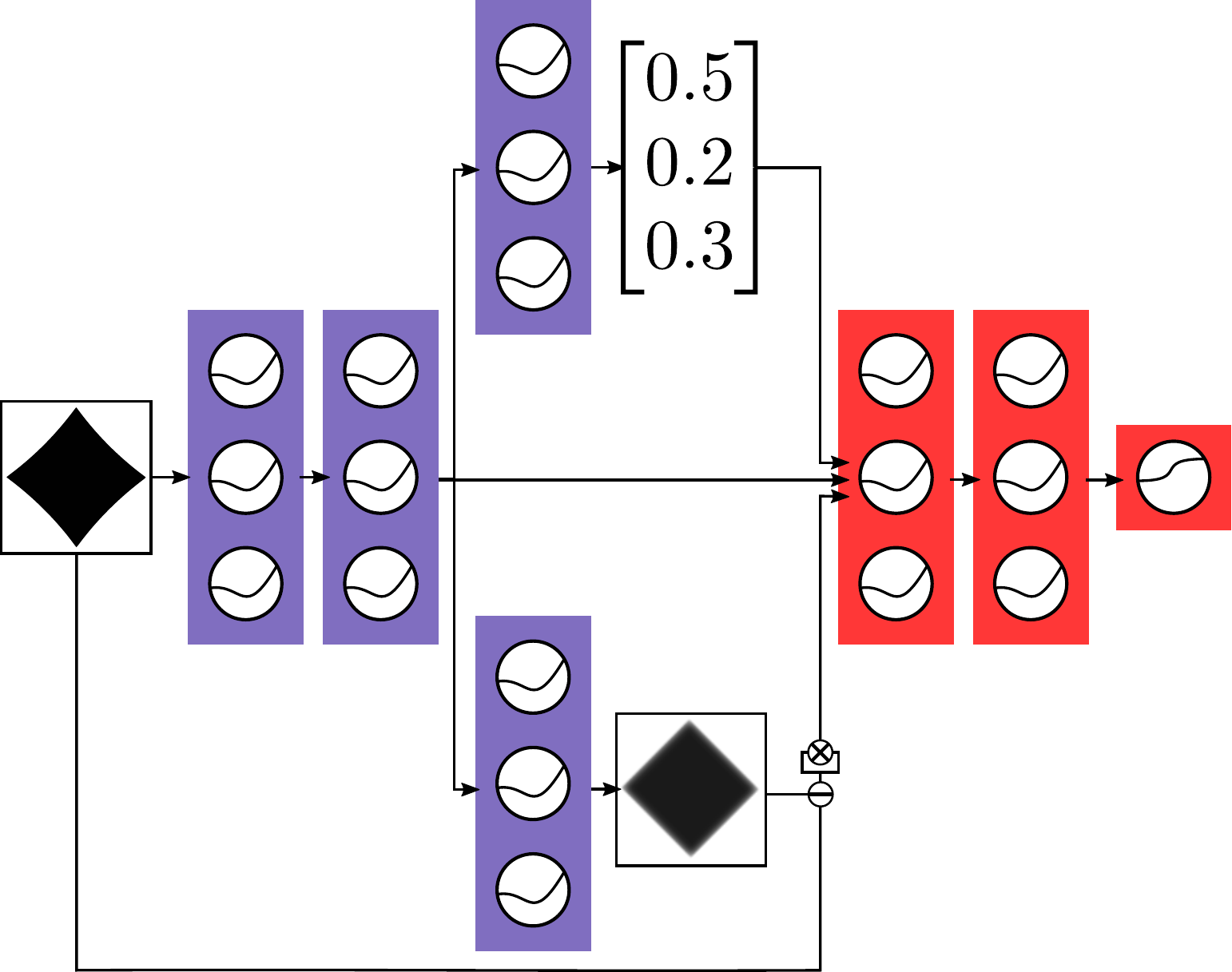}}
	\caption{A neural network classifying a diamond image with an auxiliary decoder and an abnormality module. Circles are neurons, either having a GELU or sigmoid activation. The blurred diamond reconstruction precedes subtraction and elementwise squaring. The probability vector is the softmax probability vector. Blue layers train on in-distribution data, and red layers train on both in- and out-of-distribution examples.}\label{fig:decoder}
\end{figure}

Having seen that softmax prediction probabilities enable abnormality detection, we now show there is other information sometimes more useful for detection. To demonstrate this, we exploit the learned internal representations of neural networks. We start by training a normal classifier and append an auxiliary decoder which reconstructs the input, shown in Figure \ref{fig:decoder}. Auxiliary decoders are sometimes known to increase classification performance \citep{swwae}. The decoder and scorer are trained jointly on in-distribution examples. Thereafter, the blue layers in Figure \ref{fig:decoder} are frozen. Then we train red layers on clean and noised training examples, and the sigmoid output of the red layers scores how normal the input is. Consequently, noised examples are in the abnormal class, clean examples are of the normal class, and the sigmoid is trained to output to which class an input belongs. After training we consequently have a normal classifier, an auxiliary decoder, and what we call an \textbf{abnormality module}. The gains from the abnormality module demonstrate there are possible research avenues for outperforming the baseline.

\subsection{TIMIT}
We test the abnormality module by revisiting the TIMIT task with a different architecture and show how these auxiliary components can greatly improve detection. The system is a three-layer, 1024-neuron wide classifier with an auxiliary decoder and abnormality module. This network takes as input 11 frames and must predict the phone of the center frame, 26 features per frame. Weights are initialized according to \citep{init}. This network trains for 20 epochs, and the abnormality module trains for two. The abnormality module sees clean examples and, as negative examples, TIMIT examples distorted with either white noise, brown noise (noise with its spectral density proportional to $1/f^2$), or pink noise (noise with its spectral density proportional to $1/f$) at various volumes. 

We note that the abnormality module is \emph{not} trained on the same type of noise added to the test examples. Nonetheless, Table \ref{tab:frametimit} shows that simple noised examples translate to effective detection of realistically distorted audio. We detect abnormal examples by comparing the typical abnormality module outputs for clean examples with the outputs for the distorted examples. The noises are from Aurora-2 and are added to TIMIT examples with 30\% volume. We also use the THCHS-30 dataset for Chinese speech. Unlike before, we use the THCHS-30 training examples rather than test set examples because fully connected networks can evaluate the whole training set sufficiently quickly. It is worth mentioning that \emph{fully connected} deep neural networks are noise robust \citep{seltzer}, yet the abnormality module can still detect whether an example is out-of-distribution. To see why this is remarkable, note that the network's frame classification error is 29.69\% on the \emph{entire} test (not core) dataset, and the average classification error for distorted examples is 30.43\%---this is unlike the bidirectional LSTM which had a more pronounced performance decline. Because the classification degradation was only slight, the softmax statistics alone did not provide useful out-of-distribution detection. In contrast, the abnormality module provided scores which allowed the detection of different-but-similar examples. In practice, it may be important to determine whether an example is out-of-distribution even if it does not greatly confuse the network, and the abnormality module facilitates this.

\begin{table}
\begin{center}
\begin{tabularx}{\textwidth}{X | *{6}{>{\hsize=.45\hsize}Y}}
\hline In-Distribution /\newline Out-of-Distribution & AUROC\newline/Base Softmax & AUROC\newline/Base AbMod & AUPR In/Base Softmax & AUPR In/Base AbMod & AUPR Out/Base Softmax & AUPR Out/Base AbMod\\ \cline{1-7}
\bf{TIMIT/+Airport}	& 75\scalebox{1.}{/50} & 100\scalebox{1.}{/50}
& 77\scalebox{1.}{/41} & 100\scalebox{1.}{/41} &
73\scalebox{1.}{/59} & 100\scalebox{1.}{/59} \\
\bf{TIMIT/+Babble}	& 94\scalebox{1.}{/50} & 100\scalebox{1.}{/50}
& 95\scalebox{1.}{/41} & 100\scalebox{1.}{/41} &
91\scalebox{1.}{/59} & 100\scalebox{1.}{/59} \\
\bf{TIMIT/+Car}		& 70\scalebox{1.}{/50} & 98\scalebox{1.}{/50}
& 69\scalebox{1.}{/41} & 98\scalebox{1.}{/41} &
70\scalebox{1.}{/59} & 98\scalebox{1.}{/59} \\
\bf{TIMIT/+Exhib.}	& 91\scalebox{1.}{/50} & 98\scalebox{1.}{/50}
& 92\scalebox{1.}{/41} & 98\scalebox{1.}{/41} &
91\scalebox{1.}{/59} & 98\scalebox{1.}{/59} \\
\bf{TIMIT/+Rest.}	& 68\scalebox{1.}{/50} & 95\scalebox{1.}{/50}
& 70\scalebox{1.}{/41} & 96\scalebox{1.}{/41} &
67\scalebox{1.}{/59} & 95\scalebox{1.}{/59} \\
\bf{TIMIT/+Subway}	& 76\scalebox{1.}{/50} & 96\scalebox{1.}{/50}
& 77\scalebox{1.}{/41} & 96\scalebox{1.}{/41} &
74\scalebox{1.}{/59} & 96\scalebox{1.}{/59} \\
\bf{TIMIT/+Street}	& 89\scalebox{1.}{/50} & 98\scalebox{1.}{/50}
& 91\scalebox{1.}{/41} & 99\scalebox{1.}{/41} &
85\scalebox{1.}{/59} & 98\scalebox{1.}{/59} \\
\bf{TIMIT/+Train}	& 80\scalebox{1.}{/50} & 100\scalebox{1.}{/50}
& 82\scalebox{1.}{/41} & 100\scalebox{1.}{/41} &
77\scalebox{1.}{/59} & 100\scalebox{1.}{/59} \\
\bf{TIMIT/Chinese}	& 79\scalebox{1.}{/50} & 90\scalebox{1.}{/50}
& 41\scalebox{1.}{/12} & 66\scalebox{1.}{/12} &
96\scalebox{1.}{/88} & 98\scalebox{1.}{/88} \\
\hline
Average & 80 & 97 & 77 & 95 & 80 & 98 \\
\hline
\end{tabularx}
\caption{Abnormality modules can generalize to novel distortions and detect out-of-distribution examples even when they do not severely degrade accuracy. All values are percentages.}\label{tab:frametimit}
\end{center}
\end{table}

\subsection{MNIST}
Finally, much like in a previous experiment, we train an MNIST classifier with three layers of width 256. This time, we also use an auxiliary decoder and abnormality module rather than relying on only softmax statistics. For abnormal examples we blur, rotate, or add Gaussian noise to training images. Gains from the abnormality module are shown in Table \ref{tab:mnistround2}, and there is a consistent out-of-sample detection improvement compared to softmax prediction probabilities. Even for highly dissimilar examples the abnormality module can further improve detection.

\begin{table}
\begin{center}
\begin{tabularx}{\textwidth}{X | *{6}{>{\hsize=.36\hsize}Y}}
\hline In-Distribution /\newline Out-of-Distribution & AUROC\newline/Base Softmax & AUROC\newline/Base AbMod & AUPR In/Base Softmax & AUPR In/Base AbMod & AUPR Out/Base Softmax & AUPR Out/Base AbMod \\ \cline{1-7}
\bf{MNIST/Omniglot}				& 95\scalebox{1.}{/50} & 100\scalebox{1.}{/50} & 
95\scalebox{1.}{/52} & 100\scalebox{1.}{/52} &
95\scalebox{1.}{/48} & 100\scalebox{1.}{/48} \\
\bf{MNIST/notMNIST}				& 87\scalebox{1.}{/50} & 100\scalebox{1.}{/50} &
88\scalebox{1.}{/50} & 100\scalebox{1.}{/50} &
90\scalebox{1.}{/50} & 100\scalebox{1.}{/50}  \\
\bf{MNIST/CIFAR-10bw}			& 98\scalebox{1.}{/50} & 100\scalebox{1.}{/50} &
98\scalebox{1.}{/50} & 100\scalebox{1.}{/50} &
98\scalebox{1.}{/50} & 100\scalebox{1.}{/50} \\
\bf{MNIST/Gaussian}				& 88\scalebox{1.}{/50} & 100\scalebox{1.}{/50} &
88\scalebox{1.}{/50} & 100\scalebox{1.}{/50} &
90\scalebox{1.}{/50} & 100\scalebox{1.}{/50} \\
\bf{MNIST/Uniform}				& 99\scalebox{1.}{/50} & 100\scalebox{1.}{/50} &
99\scalebox{1.}{/50} & 100\scalebox{1.}{/50} &
99\scalebox{1.}{/50} & 100\scalebox{1.}{/50} \\
\hline
Average & 93 & 100 & 94 & 100 & 94 & 100 \\
\hline
\end{tabularx}
\caption{Improved detection using the abnormality module. All values are percentages.}\label{tab:mnistround2}
\end{center}
\end{table}

\section{Discussion and Future Work}
The abnormality module demonstrates that in some cases the baseline can be beaten by exploiting the representations of a network, suggesting myriad research directions. Some promising future avenues may utilize the intra-class variance: if the distance from an example to another of the same predicted class is abnormally high, it may be out-of-distribution \citep{giryes}. Another path is to feed in a vector summarizing a layer's activations into an RNN, one vector for each layer. The RNN may determine that the activation patterns are abnormal for out-of-distribution examples. Others could make the detections fine-grained: is the out-of-distribution example a known-unknown or an unknown-unknown? A different avenue is not just to detect correct classifications but to output the probability of a correct detection. 
These are but a few ideas for improving error and out-of-distribution detection.

We hope that any new detection methods are tested on a variety of tasks and architectures of the researcher's choice. A basic demonstration could include the following datasets: MNIST, CIFAR, IMDB, and tweets because vision-only demonstrations may not transfer well to other architectures and datasets. Reporting the AUPR and AUROC values is important, and so is the underlying classifier's accuracy since an always-wrong classifier gets a maximum AUPR for error detection if error is the positive class. Also, future research need not use the exact values from this paper for comparisons. Machine learning systems evolve, so tethering the evaluations to the exact architectures and datasets in this paper is needless. Instead, one could simply choose a variety of datasets and architectures possibly like those above and compare their detection method with a detector based on the softmax prediction probabilities from their classifiers. These are our basic recommendations for others who try to surpass the baseline on this underexplored challenge.

\section{Conclusion}
We demonstrated a softmax prediction probability baseline for error and out-of-distribution detection across several architectures and numerous datasets. We then presented the abnormality module, which provided superior scores for discriminating between normal and abnormal examples on tested cases. The abnormality module demonstrates that the baseline can be beaten in some cases, and this implies there is room for future research. Our hope is that other researchers investigate architectures which make predictions in view of abnormality estimates, and that others pursue more reliable methods for detecting errors and out-of-distribution inputs because knowing when a machine learning system fails strikes us as highly important.

\section*{Acknowledgments}
We would like to thank John Wieting, Hao Tang, Karen Livescu, Greg Shakhnarovich, and our reviewers for their suggestions. We would also like to thank NVIDIA Corporation for donating several TITAN X GPUs used in this research.

\bibliographystyle{iclr2017_conference}
\bibliography{bibliography}

\begin{thebibliography}{41}
\providecommand{\natexlab}[1]{#1}
\providecommand{\url}[1]{\texttt{#1}}
\expandafter\ifx\csname urlstyle\endcsname\relax
  \providecommand{\doi}[1]{doi: #1}\else
  \providecommand{\doi}{doi: \begingroup \urlstyle{rm}\Url}\fi

\bibitem[Amodei et~al.(2016)Amodei, Olah, Steinhardt, Christiano, Schulman, and
  Man\'e]{colah}
Dario Amodei, Chris Olah, Jacob Steinhardt, Paul Christiano, John Schulman, and
  Dan Man\'e.
\newblock Concrete problems in ai safety.
\newblock \emph{arXiv}, 2016.

\bibitem[Bies et~al.(2012)Bies, Mott, Warner, and Kulick]{engwebtb}
Ann Bies, Justin Mott, Colin Warner, and Seth Kulick.
\newblock English Web Treebank, 2012.

\bibitem[Bulatov(2011)]{notmnist}
Yaroslav Bulatov.
\newblock {notMNIST} dataset.
\newblock 2011.

\bibitem[Davis \& Goadrich(2006)Davis and Goadrich]{auroc}
Jesse Davis and Mark Goadrich.
\newblock The relationship between precision-recall and {ROC} curves.
\newblock In \emph{International Conference on Machine Learning (ICML)}, 2006.

\bibitem[Fawcett(2005)]{ROC}
Tom Fawcett.
\newblock \emph{An introduction to ROC analysis}.
\newblock Pattern Recognition Letters, 2005.

\bibitem[Garofolo et~al.(1993)Garofolo, Lamel, Fisher, Fiscus, Pallett,
  Dahlgren, and Zue]{timit}
John Garofolo, Lori Lamel, William Fisher, Jonathan Fiscus, David Pallett,
  Nancy Dahlgren, and Victor Zue.
\newblock \emph{TIMIT Acoustic-Phonetic Continuous Speech Corpus}.
\newblock Linguistic Data Consortium, 1993.

\bibitem[Gimpel et~al.(2011)Gimpel, Schneider, O$'$Connor, Das, Mills,
  Eisenstein, Heilman, Yogatama, Flanigan, and Smith]{gimpel}
Kevin Gimpel, Nathan Schneider, Brendan O$'$Connor, Dipanjan Das, Daniel Mills,
  Jacob Eisenstein, Michael Heilman, Dani Yogatama, Jeffrey Flanigan, and
  Noah~A. Smith.
\newblock \emph{Part-of-Speech Tagging for Twitter: Annotation, Features, and
  Experiments}.
\newblock Association for Computational Linguistics (ACL), 2011.

\bibitem[Giryes et~al.(2015)Giryes, Sapiro, and Bronstein]{giryes}
Raja Giryes, Guillermo Sapiro, and Alex~M. Bronstein.
\newblock Deep neural networks with random gaussian weights: A universal
  classification strategy?
\newblock \emph{arXiv}, 2015.

\bibitem[Goodfellow et~al.(2015)Goodfellow, Shlens, and Szegedy]{goodfellow}
Ian~J. Goodfellow, Jonathon Shlens, and Christian Szegedy.
\newblock Explaining and harnessing adversarial examples.
\newblock In \emph{International Conference on Learning Representations
  (ICLR)}, 2015.

\bibitem[Graves et~al.(2006)Graves, Fern\'andez, Gomez, and Schmidhuber]{ctc}
Alex Graves, Santiago Fern\'andez, Faustino Gomez, and J\"{u}rgen Schmidhuber.
\newblock Connectionist temporal classification: Labeling unsegmented sequence
  data with recurrent neural networks.
\newblock In \emph{International Conference on Machine Learning (ICML)}, 2006.

\bibitem[Hendrycks \& Gimpel(2016{\natexlab{a}})Hendrycks and
  Gimpel]{detectadversarial}
Dan Hendrycks and Kevin Gimpel.
\newblock Methods for detecting adversarial images and a colorful saliency map.
\newblock \emph{arXiv}, 2016{\natexlab{a}}.

\bibitem[Hendrycks \& Gimpel(2016{\natexlab{b}})Hendrycks and Gimpel]{gelu}
Dan Hendrycks and Kevin Gimpel.
\newblock Bridging nonlinearities and stochastic regularizers with {G}aussian
  error linear units.
\newblock \emph{arXiv}, 2016{\natexlab{b}}.

\bibitem[Hendrycks \& Gimpel(2016{\natexlab{c}})Hendrycks and Gimpel]{init}
Dan Hendrycks and Kevin Gimpel.
\newblock Adjusting for dropout variance in batch normalization and weight
  initialization.
\newblock \emph{arXiv}, 2016{\natexlab{c}}.

\bibitem[Hirsch \& Pearce(2000)Hirsch and Pearce]{aurora}
Hans-G\"{u}nter Hirsch and David Pearce.
\newblock The {Aurora} experimental framework for the performance evaluation of
  speech recognition systems under noisy conditions.
\newblock \emph{ISCA ITRW ASR2000}, 2000.

\bibitem[Hochreiter \& Schmidhuber(1997)Hochreiter and Schmidhuber]{lstm}
Sepp Hochreiter and J{\"u}rgen Schmidhuber.
\newblock \emph{Long short-term memory}.
\newblock Neural Computation, 1997.

\bibitem[Hu \& Liu(2004)Hu and Liu]{CR}
Minqing Hu and Bing Liu.
\newblock \emph{Mining and Summarizing Customer Reviews}.
\newblock Knowledge Discovery and Data Mining (KDD), 2004.

\bibitem[Iyyer et~al.(2015)Iyyer, Manjunatha, Boyd-Graber, and Iii.]{DAN}
Mohit Iyyer, Varun Manjunatha, Jordan Boyd-Graber, and Hal~Daum\'e Iii.
\newblock \emph{Deep Unordered Composition Rivals Syntactic Methods for Text
  Classification}.
\newblock Association for Computational Linguistics (ACL), 2015.

\bibitem[Joulin et~al.(2016)Joulin, Grave, Bojanowski, and Mikolov]{fasttext}
Armand Joulin, Edouard Grave, Piotr Bojanowski, and Tomas Mikolov.
\newblock Bag of tricks for efficient text classification.
\newblock \emph{arXiv}, 2016.

\bibitem[Kingma \& Ba(2015)Kingma and Ba]{adam}
Diederik Kingma and Jimmy Ba.
\newblock \emph{Adam: A Method for Stochastic Optimization}.
\newblock International Conference for Learning Representations (ICLR), 2015.

\bibitem[Krizhevsky(2009)]{cifar}
Alex Krizhevsky.
\newblock Learning Multiple Layers of Features from Tiny Images, 2009.

\bibitem[Lake et~al.(2015)Lake, Salakhutdinov, and Tenenbaum]{omniglot}
Brenden~M. Lake, Ruslan Salakhutdinov, and Joshua~B. Tenenbaum.
\newblock Human-level concept learning through probabilistic program induction.
\newblock \emph{Science}, 2015.

\bibitem[Lang(1995)]{newsgroups}
Ken Lang.
\newblock Newsweeder: Learning to filter netnews.
\newblock In \emph{International Conference on Machine Learning (ICML)}, 1995.

\bibitem[Lewis et~al.(2004)Lewis, Yang, Rose, and Li]{reuters}
David~D. Lewis, Yiming Yang, Tony~G. Rose, and Fan Li.
\newblock Rcv1: A new benchmark collection for text categorization research.
\newblock \emph{Journal of Machine Learning Research (JMLR)}, 2004.

\bibitem[Loshchilov \& Hutter(2016)Loshchilov and Hutter]{sgdr}
Ilya Loshchilov and Frank Hutter.
\newblock Sgdr: Stochastic gradient descent with restarts.
\newblock \emph{arXiv}, 2016.

\bibitem[Maas et~al.(2011)Maas, Daly, Pham, Huang, Ng, and Potts]{maas}
Andrew~L. Maas, Raymond~E. Daly, Peter~T. Pham, Dan Huang, Andrew~Y. Ng, and
  Christopher Potts.
\newblock Learning word vectors for sentiment analysis.
\newblock In \emph{Association for Computational Linguistics (ACL)}, 2011.

\bibitem[Manning \& Sch\"{u}tze(1999)Manning and Sch\"{u}tze]{manning}
Chris Manning and Hinrich Sch\"{u}tze.
\newblock \emph{Foundations of Statistical Natural Language Processing}.
\newblock MIT Press, 1999.

\bibitem[Marcus et~al.(1993)Marcus, Marcinkiewicz, and Santorini]{WSJ}
Mitchell~P. Marcus, Mary~Ann Marcinkiewicz, and Beatrice Santorini.
\newblock Building a large annotated corpus of {E}nglish: The {P}enn
  {T}reebank.
\newblock \emph{Computational linguistics}, 1993.

\bibitem[Nguyen et~al.(2015)Nguyen, Yosinski, and Clune]{foolers}
Anh Nguyen, Jason Yosinski, and Jeff Clune.
\newblock Deep neural networks are easily fooled: High confidence predictions
  for unrecognizable images.
\newblock In \emph{Computer Vision and Pattern Recognition (CVPR)}, 2015.

\bibitem[Nguyen \& O'Connor(2015)Nguyen and O'Connor]{nguyen}
Khanh Nguyen and Brendan O'Connor.
\newblock Posterior calibration and exploratory analysis for natural language
  processing models.
\newblock In \emph{Empirical Methods in Natural Language Processing (EMNLP)},
  2015.

\bibitem[Owoputi et~al.(2013)Owoputi, O'Connor, Dyer, Gimpel, Schneider, and
  Smith]{owoputi}
Olutobi Owoputi, Brendan O'Connor, Chris Dyer, Kevin Gimpel, Nathan Schneider,
  and Noah~A. Smith.
\newblock Improved part-of-speech tagging for online conversational text with
  word clusters.
\newblock In \emph{North American Chapter of the Association for Computational
  Linguistics (NAACL)}, 2013.

\bibitem[Pang et~al.(2002)Pang, Lee, and Vaithyanathan]{MR}
Bo~Pang, Lillian Lee, and Shivakumar Vaithyanathan.
\newblock Thumbs up? sentiment classification using machine learning
  techniques.
\newblock In \emph{Empirical Methods in Natural Language Processing (EMNLP)},
  2002.

\bibitem[Provost et~al.(1998)Provost, Fawcett, and Kohavi]{provost}
Foster Provost, Tom Fawcett, and Ron Kohavi.
\newblock The case against accuracy estimation for comparing induction
  algorithms.
\newblock In \emph{International Conference on Machine Learning (ICML)}, 1998.

\bibitem[Saito \& Rehmsmeier(2015)Saito and Rehmsmeier]{auprbaseline}
Takaya Saito and Marc Rehmsmeier.
\newblock The precision-recall plot is more informative than the {ROC} plot
  when evaluating binary classifiers on imbalanced datasets.
\newblock In \emph{PLoS ONE}. 2015.

\bibitem[Seltzer et~al.(2013)Seltzer, Yu, and Wang]{seltzer}
Michael~L. Seltzer, Dong Yu, and Yongqiang Wang.
\newblock Investigation of deep neural networks for noise robust speech
  recognition.
\newblock In \emph{IEEE International Conference on Acoustics, Speech, and
  Signal Processing (ICASSP)}, 2013.

\bibitem[Steinhardt \& Liang(2016)Steinhardt and Liang]{jacob}
Jacob Steinhardt and Percy Liang.
\newblock Unsupervised risk estimation using only conditional independence
  structure.
\newblock In \emph{Neural Information Processing Systems (NIPS)}, 2016.

\bibitem[Wang \& Zhang(2015)Wang and Zhang]{chinese}
Dong Wang and Xuewei Zhang.
\newblock Thchs-30 : A free chinese speech corpus.
\newblock In \emph{Technical Report}, 2015.

\bibitem[Williams \& Renals(1997)Williams and Renals]{williams}
Gethin Williams and Steve Renals.
\newblock Confidence measures for hybrid hmm/ann speech recognition.
\newblock In \emph{Proceedings of EuroSpeech}, 1997.

\bibitem[Xiao et~al.(2010)Xiao, Hays, Ehinger, Oliva, and Torralba]{SUN}
Jianxiong Xiao, James Hays, Krista~A. Ehinger, Aude Oliva, and Antonio
  Torralba.
\newblock Sun database: Large-scale scene recognition from abbey to zoo.
\newblock In \emph{IEEE Conference on Computer Vision and Pattern Recognition
  (CVPR)}, 2010.

\bibitem[Yu et~al.(2010)Yu, Li, and Deng]{yu}
Dong Yu, Jinyu Li, and Li~Deng.
\newblock Calibration of confidence measures in speech recognition.
\newblock In \emph{IEEE Transactions on Audio, Speech, and Language}, 2010.

\bibitem[Zagoruyko \& Komodakis(2016)Zagoruyko and Komodakis]{wrn}
Sergey Zagoruyko and Nikos Komodakis.
\newblock Wide residual networks.
\newblock \emph{British Machine Vision Conference}, 2016.

\bibitem[Zhang et~al.(2016)Zhang, Lee, and Lee]{swwae}
Yuting Zhang, Kibok Lee, and Honglak Lee.
\newblock Augmenting supervised neural networks with unsupervised objectives
  for large-scale image classification.
\newblock In \emph{International Conference on Machine Learning (ICML)}, 2016.

\end{thebibliography}

\newpage
\appendix
\section{Abnormality Module Example}

\end{document}